# Gesture Controlled Robot For Human Detection

*Athira T.S [a], Honey Manoj [a], R S Vishnu Priya [a], Vishnu K Menon [a], Srilekshmi M [b]*

[a] *Department of Electronics and Communication, SCMS School of Engineering and Technology, Karukutty ,Ernakulam-683576, India*
[b] *Department of Electronics and Communication, SCMS School of Engineering and Technology, Karukutty ,Ernakulam-683576, India*

A B S T R A C T

It is very important to locate survivors from collapsed buildings so that rescue operations can be arranged. Many lives are lost due to lack of competent systems to detect people in these collapsed buildings at the right time. So here we have designed a hand gesture controlled robot which is capable of detecting humans under these collapsed building parts. The proposed work can be used to access specific locations that are not humanly possible, and detect those humans trapped under the rubble of collapsed buildings. This information is then used to notify the rescue team to take adequate measures and initiate rescue operations accordingly.

## 1. Introduction

The rate of manifestation of disasters has increased drastically which has caused widespread loss of human life. Disasters could be natural such as earthquakes or cyclones and human-induced such as industrial accidents, gas explosions etc. One of the most common disasters that causes casualties in urban cities is an earthquake.

It has been observed that the number of earthquakes is increasing. It has been reported that this increase in earthquakes could be due to the intense activity of the tectonic plates. An earthquake causes major injuries and loss of many lives. Roads, bridges and property damage is observed, and it also causes collapsing or destabilization of buildings. The collapsing of buildings causes many people to be trapped in the debris. It is very important that these people have to be rescued at the right time so that the loss of life is prevented.

When an Earthquake strikes, most casualties are due to collapse of buildings, as the population density is high in cities the number of casualties is likely to increase. People often get trapped under the rubble of the collapsed building. A healthy uninjured human being can survive up to 72 hours with sufficient air supply, but as we cannot predict the situation of people trapped inside the collapsed building therefore rescue operation should be commenced as soon as possible. The rescue operation will not be effortless as there might be a situation where a person is trapped in a location that is not humanly accessible. Hence in the proposed project 'Gesture Controlled Robot for Human Detection' which would be used to access specific locations that are not humanly possible to detect victims that are trapped under the rubble of collapsed buildings and notify about the situation to the rescue team so that they could take adequate measures and initiate rescue operations accordingly. It has two units, the control unit and the robot. The control section is used to provide the hand gestures [16] to the robotic section for its movement. The robotic section moves with the help of these hand gestures and also provides human detection. Human detection is done using a PIR sensor. When a human is detected by the PIR sensor this information is transmitted back using the transceiver to the control section to initiate rescue operations.

*R S Vishnu Priya. Tel.: 9048076999
E-mail address: vpriyasajeevkumar@gmail.com

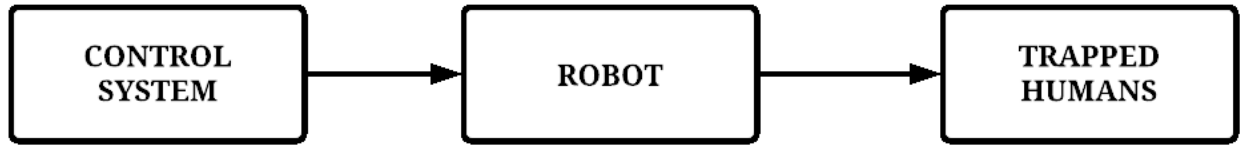

**Figure 1.1. General scenario**

## 2. Related Works

The following are works related to the proposed system.

Di Zhang et al. [1] proposed a sensor-based system to detect humans trapped under the debris. In this paper, a carbon dioxide sensor, an oxygen sensor, a thermal camera, and microphone was used. Results conveyed that oxygen measurements were not useful to detect human presence. The efficiency of carbon dioxide and the thermal camera were only 66.67% and 77.78% respectively. Another limitation was the difficulty in using the gas sensor in open areas due to strong airflow which affects the carbon dioxide levels. Also this system is not cost effective.

Prof. Chethan Bulla et al. [2] proposed a gesture-controlled robot that uses image processing to identify the object and change street light intensity according to the size of the object detected. Benjula Anbu Malar et al. [3] suggested a hand gesture-controlled robot using a 3-axis accelerometer but didn't specify a particular application for it.

Lavanya K N et al. [4] proposed a robot in which the movement was controlled using hand gestures. The methodology included three steps - capturing of the image, processing of the image, and then extraction of data from the image.

Ankita Saxena et al. [5] also suggested a gesture-controlled robot using image processing. It used the artificial neural network to identify the various gestures. Image processing was done that involved training the model which was difficult and required bulk data.

S Mohith et al. [6] proposed a robot that controlled using both hand gestures and voice control. The system also had an obstacle detector. The use of both gesture and voice control is complicated.

Asha Gupta et al. [7] suggested a human detection robot that uses an ATMEGA16 microcontroller, a temperature sensor, and a PIR sensor. In this paper, PIR sensor and temperature sensor were used to detect the presence of humans. The limitations of this system are that the temperature sensor fails to detect human presence in case of a fire and the PIR sensor cannot detect a stationary object.

Vishakha Borkar et al. [8] designed a revolutionary microwave-based life detection system using L band to detect humans trapped under earthquake rubble. The system detects the breathing and heartbeat signals of trapped human beings. Signal processing of these signals was used in determining the status of the person. This system is expensive and the L band frequency cannot penetrate metal-like structures. If a clutter signal is involved, it may destroy vital information about life which was a disadvantage of this system.

Apurva S Ubhale et al. [9] developed a microwave life-detection system to identify humans trapped under the debris. The system consists of a microwave frequency electromagnetic signal to detect whether the human is in motion or not. The operation principle of the motion detection system is based on the Doppler frequency shift of the wave. Once motion is detected using Doppler radar, then to identify if the object is a human or not, for this purpose the microwave test bench is used. The problem with Doppler radar is that it requires specific calibration and even minute changes in the environmental conditions can result in an impairment of the performance of the system. This system has a narrow angle view which makes it inappropriate for wide disaster areas.

Bethanney Janney et al. [10] developed a human detection robot that consists of a piezoelectric plate. This piezoelectric sensor is used to sense the minute vibrations of the human under debris. It then collects the data and sends the data to the collection unit.

In the proposed work, a new and unique method for the detection of humans trapped under collapsed buildings is proposed. The proposed system is also capable of detecting stationary animals which is an added advantage.

## 3. Proposed Work

Today robots play a very crucial role in almost all sectors of society. The proposed work is a gesture-controlled robot which uses Arduino Uno. Hand gestures provide a more schematic way of controlling the robot. Hand gestures are natural, and by using wireless communication we can easily interact with the robot in a more friendly way. The robot's movement depends on the hand gestures.

Hand gestures result in a change in the x and y coordinates of the accelerometer which helps in the movement of the robot. For this purpose, we have used an accelerometer. The gesture-controlled robot can be used to detect humans trapped under the collapsed buildings. Human detection is done using a PIR sensor. When a human is detected by the PIR sensor this information is transmitted back using the transceiver to the control section to initiate rescue operations.

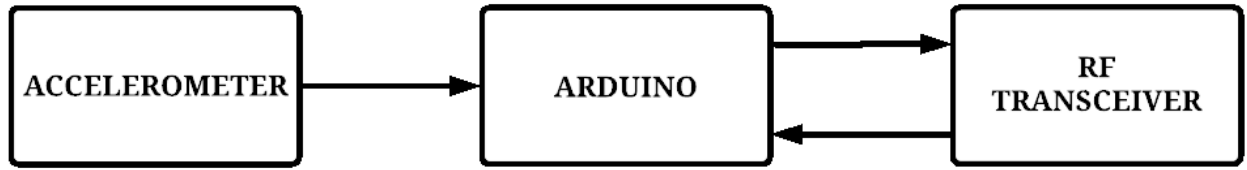

Figure 3.1. Block diagram of Control Unit

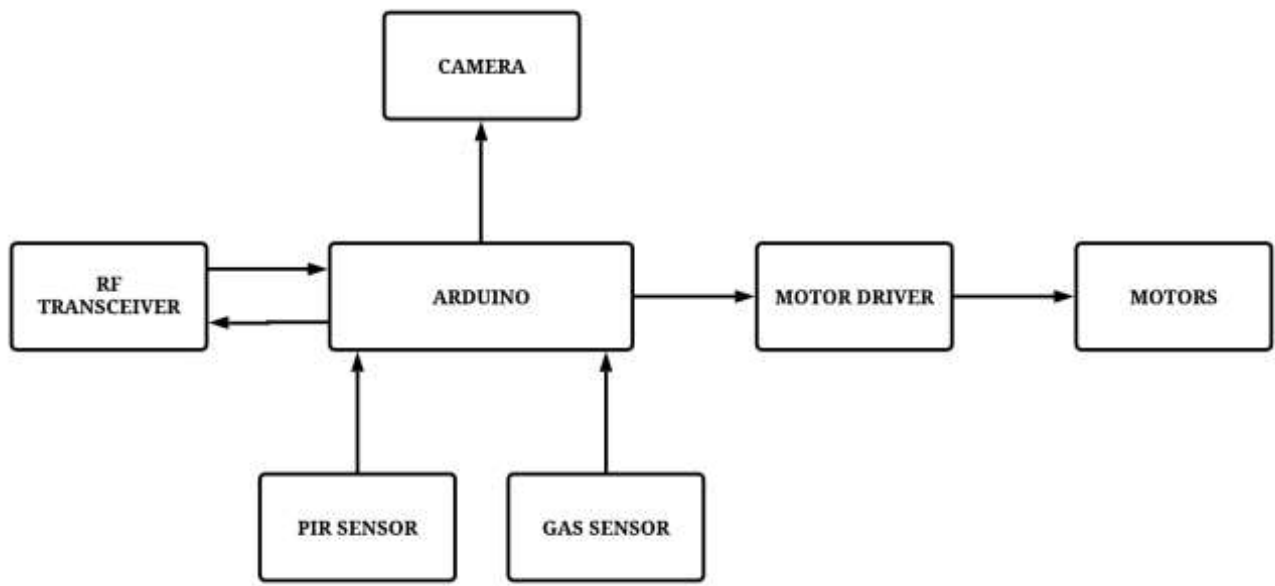

Figure 3.2. Block diagram of Human Detection System

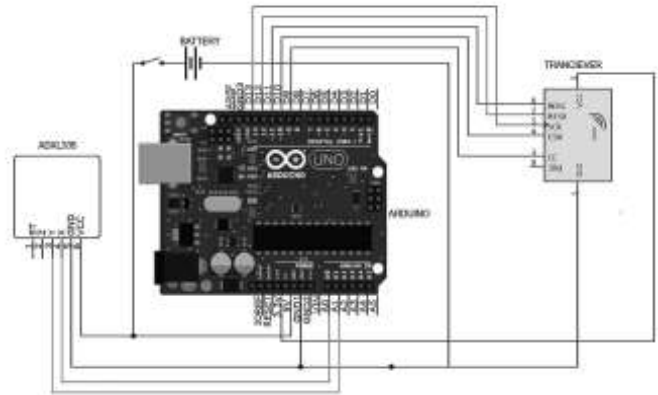

**Figure 3.3. Circuit diagram of Control Unit**

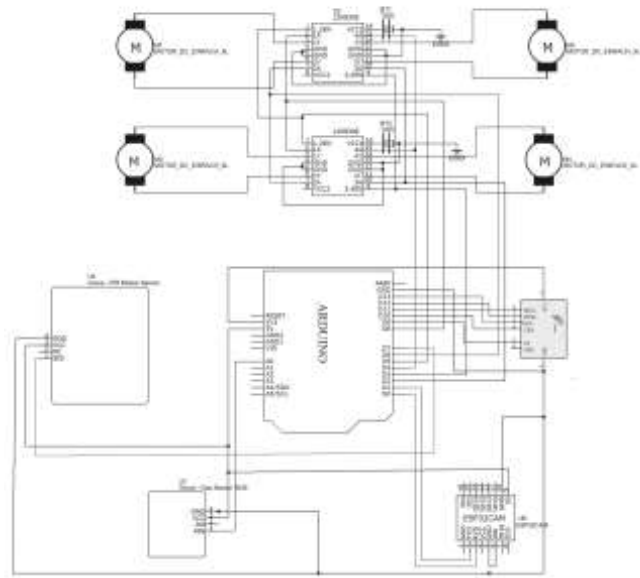

**Figure 3.4. Circuit diagram of Human Detection System**

*3.1 Working*

The system consists of two units: the control unit and the robot. The components used in the control section are Arduino Uno, Accelerometer [ADXL335] [12], Transceiver [nRF24L01], and a battery. Arduino Uno has 6 analog and 14 digital pins to which the hardware components are connected. The accelerometer [ADXL335] has 6 pins. The pins namely X, Y, GND, and VCC are connected to A0, A1, GND1, and 5V pins of Arduino Uno respectively. The digital I/O pins D9, D10, D11, D12, and D13 of the Arduino are connected to the Transceiver pins CE(3), CSN(4), MOSI(6), MISO(7), and SCK(5) respectively. The VCC and GND pins of all the components are connected to the positive and negative terminals of the battery respectively.

The components used in the robot section are Arduino Uno, Transceiver [nRF24L01], PIR sensor [HCSR 501], Gas sensor [MQ-9], ESP32CAM, Motor driver [L293D] and motors. The pins namely SIG(1), VCC(3), and GND(4) of the Gas sensor [MQ-9] are connected to the Arduino pins A0, 5V, and GND1 respectively. The digital I/O pins D9, D10, D11, D12, and D13 of the Arduino are connected to the Transceiver pins CE(3), CSN(4), MOSI(6), MISO(7), and SCK(5) respectively. The pins GND(1), TXD(2), RXS(3), and 5V(9) of ESP32CAM are connected to the Arduino pins GND, D1, D0, and 5V. The pin numbers 5 and 6 of ESP32CAM are shorted. The pins SIG(1) and VCC(3) of the PIR sensor are connected to the Arduino pins D7 and 5V. Pin number 4 of the PIR sensor is connected to the common ground. The pins 1,2EN(1), 1A(2), 2A(7), 3,4EN(9), 3A(10), and 4A(15) of Motor driver [L293D] are connected to the Arduino pins D5, D8, D6, D3, D2, and D4 respectively. Motors are connected to the output pins of Motor driver [L293D]. The battery's positive terminal is connected to pin number 16 of Motor driver [L293D] and all the ground pins are connected to the negative terminal.

The module consists of two parts, the first part consisting of the accelerometer [ADXL 335] [13], the Arduino Uno, and the transceiver [nRF24L01]. The accelerometer will measure the x and y coordinates of the hand [15]. These values are analog in nature and they help the robot move in forward, backward, left and right directions. These analog values are then given to the A0 and A1 pins of the Arduino. The Arduino converts these analog values into digital values and transmits this information through the transceiver which will facilitate the movement of the robotic system.

The second part consists of the transceiver [nRF24L01], Arduino Uno, motor driver [L293D] [14], motors, PIR sensor [HCSR 501], Gas sensor [MQ-9], and an ESP32 camera. The transceiver receives the information of the x and y coordinates detected using the accelerometer at the control unit and sends this information to the motor driver through the Arduino and initiates the movement of the robotic system in forward, backward, right or left directions.

The section also consists of a human detection system. The Arduino Uno takes input from the PIR sensor [HCSR 501] if a human is detected. The PIR sensor is given a constant oscillatory motion for the detection of stationary humans since they may be trapped under the debris. A gas sensor [MQ-9] is also provided for sensing the presence of Carbon Monoxide, LPG gas, and Methane. The information from the gas sensor is given to the Arduino, the Arduino then transmits this information back using the transceiver, this helps the people being involved in the rescue operation to take protective measures. A ESP32 camera is provided for real-time monitoring to facilitate the easy movement of the robot. Once a human is detected by the PIR sensor this information is given to the Arduino, the Arduino then transmits this information back to the transceiver at the control unit. The information, when it is received, is used to initiate rescue operations.

## 4. Result Analysis

PIR sensor is used for the detection of humans, which is given a constant motion to detect stationary humans and animals. The detected information from the PIR sensor is connected to Arduino Uno. The Arduino [11] then transmits this information to the transceiver at the control unit and the rescue operations are initiated.

The robot can be controlled from a maximum distance of 1000 meters. The transceiver which works at 2.4 GHz was used for this purpose. The coordinates of the hand was measured by the accelerometer and the information transmitted was able to control the motor driver and the required movement was observed for each hand gesture. The robot was able to identify the human presence and transmit this information to the control unit.

## 5. Conclusions

This work will be a great help to rescue teams for detecting human beings at disaster sites. The system helps to save precious lives and reduce the death rate from such disasters to a greater extent.

The system can be advanced to detect multiple people at a time. This robot can also be used in military and defence applications for the detection of enemies hiding. Through this we will be able to save a lot of people trapped at the right time and save their lives. The system has a large progressive scope in complex models for robots performing complicated tasks. The proposed work has used only a limited number of hand gestures for movement of the robot and by adding more hand gestures the performance of the robot can be improved. Due to these progressive scopes and societal needs the proposed system has great potential. Robots have a wide range of applications and they can be used in inaccessible places. With the execution of various technologies and sensors this system can be used for the purpose of detection of various metals. By giving this robot intelligence it need not be controlled by humans and can work by itself.